\documentclass[10pt,twocolumn,letterpaper]{article}

\usepackage[pagenumbers]{cvpr} 
\makeatletter
\@namedef{ver@everyshi.sty}{}
\makeatother
\usepackage{graphicx}
\usepackage{amsmath}
\usepackage[ruled,vlined]{algorithm2e}

\SetCommentSty{mycommfont}
\usepackage[table,xcdraw]{xcolor}
\usepackage{multirow}
\usepackage{color, soul}
\sethlcolor{red}

\newcommand{\val}[2]{#1$\pm$#2}

\SetKwInput{KwInput}{Input}                
\SetKwInput{KwOutput}{Output}              

\usepackage{pgfplots}
\usepackage{tikz}
\usetikzlibrary{shapes.multipart}
\usetikzlibrary{shapes.geometric,arrows} 
\pgfplotsset{compat=1.9}

\pgfplotsset{
    myplotstyle/.style={
    legend style={draw=none, font=\small},
    legend cell align=left,
    legend pos=north east,
    ylabel style={align=center},
    xlabel style={align=center},
    x tick label style={},
    y tick label style={},
    xtick pos=left, 
    ytick pos=left,
    scaled ticks=false,
    every axis plot/.append style={thick},
    },
}

\newcommand*{\green}[1]{
    \protect
  \begin{tikzpicture}[scale=#1]
    \protect \draw[draw=green!65, fill=green!70] (0,0) -- (1,0) -- (0.5,0.87) -- cycle;
  \end{tikzpicture}
}

\newcommand*{\red}[1]{%
    \protect
  \begin{tikzpicture}[scale=#1]
    \protect \draw[draw=red!55, fill=red!70] (0,0) -- (1,0) -- (0.5,-0.87) -- cycle;
  \end{tikzpicture}%
}

\newcommand*{\invred}[1]{%
    \protect
  \begin{tikzpicture}[scale=#1]
    \protect \draw[draw=red!55, fill=red!70] (0,0) -- (1,0) -- (0.5,0.87) -- cycle;
  \end{tikzpicture}%
}

\newcommand*{\up}[1]{\begin{centering}{\green{0.25}{\hspace{1mm}#1}}\end{centering}}
\newcommand*{\down}[1]{\begin{centering}{\red{0.25}{\hspace{1mm}#1}}\end{centering}}
\newcommand*{\invdown}[1]{\invred{0.25}{\hspace{1mm}#1}}

\newcommand*{\upp}[1]{\green{0.25}{\hspace{1mm}#1}\%}
\newcommand*{\downp}[1]{\red{0.25}{\hspace{1mm}#1}\%}

\usepackage[pagebackref,breaklinks,colorlinks]{hyperref}

\usepackage[capitalize]{cleveref}
\crefname{section}{Sec.}{Secs.}
\Crefname{section}{Section}{Sections}
\Crefname{table}{Table}{Tables}
\crefname{table}{Tab.}{Tabs.}

\def\cvprPaperID{88} 
\def\confName{CVPR}
\def\confYear{2023}

\begin{document}

\title{Can we learn better with hard samples?}

\author{Subin Sahayam, John Zakkam, Umarani Jayaraman \\
Indian Institute of Information Technolgy, Kancheepuram\\
{\tt\small \{coe18d001, ced18i059, umarani\}@iiitdm.ac.in}
}
\maketitle

\begin{abstract}
In deep learning, mini-batch training is commonly used to optimize network parameters. However, the traditional mini-batch method may not learn the under-represented samples and complex patterns in the data, leading to a longer time for generalization. To address this problem, a variant of the traditional algorithm has been proposed, which trains the network focusing on mini-batches with high loss. The study evaluates the effectiveness of the proposed training using various deep neural networks trained on three benchmark datasets (CIFAR-10, CIFAR-100, and STL-10). The deep neural networks used in the study are ResNet-18, ResNet-50, Efficient Net B4, EfficientNetV2-S, and MobilenetV3-S. The experimental results showed that the proposed method can significantly improve the test accuracy and speed up the convergence compared to the traditional mini-batch training method. Furthermore, we introduce a hyper-parameter delta ($\delta$) that decides how many mini-batches are considered for training. Experiments on various values of  $\delta$ found that the performance of the proposed method for smaller $\delta$ values generally results in similar test accuracy and faster generalization. We show that the proposed method generalizes in 26.47\% less number of epochs than the traditional mini-batch method in EfficientNet-B4 on STL-10. The proposed method also improves the test top-1 accuracy by 7.26\% in ResNet-18 on CIFAR-100.
\end{abstract}

\section{Introduction}
\label{sec:intro}

Deep Neural Networks (DNNs) over the years have stood out in many representation learning tasks. The back-propagation algorithm is the method of choice for training neural networks\cite{rumelhart1986learning, werbos1974beyond}. The back-propagation algorithm allows multi-layer neural networks to learn complex representations between the inputs and outputs \cite{hinton2006reducing, krizhevsky2017imagenet}. It overcomes the limitation of learning linearly separable vectors in neural networks like the perceptron \cite{rosenblatt1958perceptron}. Essentially, the more complex the data, the more back-propagations are required. The field of deep learning has progressed from learning simple linear representations using simple artificial neural networks to learning highly complex fine-grained representations using transformers, all using back-propagation.

\begin{figure}[t]
    \centering
    \begin{tikzpicture}
    \begin{axis}[myplotstyle, xlabel=Number of epochs, ylabel=Cross Entropy Loss, legend entries={$\delta = 1$, $\delta = 0.2$, $\delta = 0.5$, $\delta = 0.8$}, legend style={draw=none}]
    \addplot+[smooth, mark=*] table [x=iter,y=delta_1.0_resnet18_cifar10_TrainLoss, col sep=comma] {runs_128.csv};
    \addplot+[smooth, mark=*] table [x=iter, y=delta_0.2_resnet18_cifar10_TrainLoss, col sep=comma] {runs_128.csv};
    \addplot+[smooth, mark=*] table [x=iter, y=delta_0.5_resnet18_cifar10_TrainLoss, col sep=comma] {runs_128.csv};
    \addplot+[smooth, mark=*] table [x=iter, y=delta_0.8_resnet18_cifar10_TrainLoss, col sep=comma] {runs_128.csv};
    \end{axis}
    \end{tikzpicture}    
    \caption{Comparing the convergence of ResNet-18 \cite{he2016deep} with different $\delta$ values on CIFAR-10 \cite{krizhevsky2009learning}. $\delta = 1$ represents the traditional mini-batch training \cite{ruder2016overview}, other values of $\delta$ represent the ablations to the proposed method. }
    \label{fig:teaser_fig}
\end{figure}
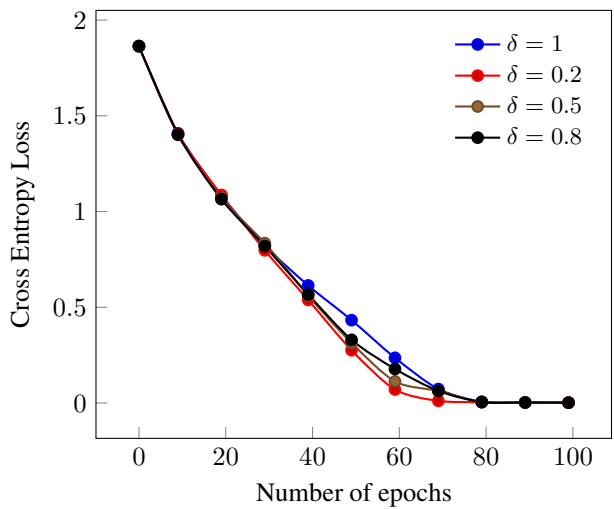



The back-propagation algorithm in neural networks can be applied in batches (Batch Gradient Descent), on every sample (Stochastic Gradient Descent) or even in mini-batches (Mini-batch Gradient Descent) \cite{ruder2016overview}. In the Batch Gradient Descent algorithm, back-propagation is done on the average of the gradients over all the samples of the dataset. It can take a lot of computation time for generalization. The Stochastic Gradient Descent (SGD) algorithm uses one sample in every iteration to compute the gradients and update the weights. However, SGD may never result in a global minimum, and the network might not converge as the gradients can get stuck at local minima \cite{pruthi2020estimating, dogo2018comparative}. Mini-batch Gradient Descent solves these problems, a mini-batch consists of a fixed number of training examples that is less than the actual dataset size. So, in each iteration, the network is trained on a different group of batches until all samples in the dataset are used. Mini-batch Gradient Descent generalizes faster than batch gradient descent and it has a lesser chance of getting stuck at local minima \cite{staib2019escaping}. Hard samples are the ones that may be under-represented in the whole dataset or might have a complex representation that might require more iterations to learn. These samples might require a higher weightage compared to the other samples in the dataset. Such samples will generally result in a higher loss value after back-propagation. One of the most popular algorithmic approaches that assign weight to hard samples is focal loss \cite{lin2017focal}. The problem with focal loss is that it has an $\alpha$ and $\gamma$ hyper-parameters which are decided before training \cite{mukhoti2020calibrating}.

While the back-propagation algorithm has enabled neural networks to learn complex representations, it remains a challenge to learn hard samples in the data \cite{bengio2009learning, lecun2015deep, arpit2017closer}. The cost of not being able to learn hard samples from the dataset leads to a slower convergence. Additionally \cite{geman1984stochastic, twomey1998bias}, neural networks tend to have reducible errors namely, bias  and variance. One well-known solution to the problem is to increase the depth of the network which improves the network's ability to generalize and learn finer and more complex latent representations \cite{neal2018modern}. Learning from hard samples in the data is essential as it could improve the performance of the trained network. From the literature, deep neural networks reduce variance and bias generally converges faster \cite{krawczyk2016learning, fernandez2014we, chawla2010data}. Neural networks in recent times have become over-parameterized to overcome limitations such as reducible errors \cite{neal2018modern}. Thus, it is important to study methods that can improve generalization in neural networks. The authors believe that even small progress toward better generalization is an important problem that would have a high impact on the field of deep learning.

In this paper, the authors propose a variation of the mini-batch training method focusing on learning hard samples in the dataset. It aims to help neural networks converge faster with minimal change in the test accuracy with respect to the traditional training method. The intuition behind the proposed method is the following observation - \textit{When preparing for an exam, students tend to spend more time focusing on difficult concepts compared to easier ones.} The proposed method introduces a new hyper-parameter $\delta$ which selects a fraction of mini-batches that are considered hard mini-batches for the next iteration in the training process. The authors define hard mini-batches as mini-batches arranged in non-increasing order of loss values. For the process of selecting a mini-batch, $\delta$ can take values from $(0, 1]$, where $1$ corresponds to the selection of all the mini-batches. For example, $\delta$ values $ 0.2, 0.5, 0.8, 1$ correspond to the selection of $20 \%, 50 \%, 80 \%$ and $100 \%$ mini-batches arranged in non-increasing order of loss values for the next training iteration. Figure \ref{fig:teaser_fig} shows that varying values of $\delta$ help in faster convergence in ResNet 18 \cite{he2016deep} on the CIFAR-10 dataset \cite{krizhevsky2009learning}. The proposed method for $\delta = 0.2$ achieves $9.58 \%$ faster convergence with the same number of back-propagations compared to the traditional training method.

\section{Related Work}
\label{sec:related_work}

\noindent \textbf{Representation Learning.}
Representation learning is an area of research in machine learning and artificial intelligence that aims to learn useful features or representations from raw data. The field has gained significant attention in recent years due to its potential to improve the performance of various machine learning tasks, including image classification, speech recognition, natural language processing, and recommender systems \cite{zhang2019deep, bengio2013representation}. One of the main challenges in representation learning is to design effective algorithms that can learn meaningful representations from high-dimensional data. Deep learning is the most popular representation learning approach that involves training deep neural networks to extract hierarchical and abstract features from raw data. Deep learning has achieved state-of-the-art results in many computationally complex tasks, such as image recognition, speech recognition, and natural language processing. AlexNet \cite{krizhevsky2017imagenet} is one of the first deep learning networks that achieved a significant breakthrough in image classification performance on the ILSVRC image classification challenge in 2012. Some of the other popular deep-learning networks for image classification that followed are VGGNet \cite{simonyan2014very}, ResNet \cite{he2016deep}, DenseNet \cite{huang2017densely}, MobileNet \cite{howard2019searching}, EfficientNet \cite{tan2019efficientnet}, Vision Transformer \cite{dosovitskiy2020image}, and Swin Transformer \cite{liu2021swin} networks.

Another important aspect of supervised representation learning is the evaluation of learned representations. It can be challenging due to the lack of a clear metric or benchmark for measuring their quality \cite{ozair2019wasserstein}. Recent work \cite{neyshabur2020being, guo2019spottune, iman2023review} have proposed to use of transfer learning, where pre-trained representations on one task are transferred to another task, as a way to evaluate the quality of learned representations. For example, \cite{frome2013devise} introduced the Deep Visual-Semantic Embedding (DeViSE) network that learned a joint embedding space for images and their associated textual descriptions and demonstrated its effectiveness on various tasks. Despite the significant progress, there are still many challenges and open questions in representation learning, such as the design of more efficient algorithms, the evaluation of learned representations, and the integration of multiple modalities. \\

\noindent \textbf{Neural Networks.}
Training deep neural networks requires extensive experimentation. One of the early breakthroughs in training neural networks is the back-propagation algorithm \cite{werbos1974beyond}, later popularized in the work \cite{rumelhart1986learning} which emphasized learning representations through back-propagation. In every step of training a neural network, there are two passes, one forward pass to predict the error on the set of samples, and one backward pass (back-propagation) to update the weights of the network according to the gradient of the error. Back-propagation has shown that over-parameterized networks such as deep convolutional neural networks, and auto-encoders can converge on the training set with minimal error. However, due to their over-parameterized nature, these models in principle have the capacity to over-fit any set of labels including pure noise. To control the rate of learning, optimizers such as SGD with momentum \cite{ruder2016overview}, Nesterov momentum \cite{botev2017nesterov} Adam \cite{kingma2014adam}, Lamb \cite{you2019large} along with learning rate schedulers such as Step Learning Rate (LR) \cite{kim2021automated}, Cosine LR \cite{loshchilov2016sgdr} is used. Learning rate, mini-batch size, and the number of iterations to train are all pre-defined hyper-parameters for the process of training. Hyper-parameter tuning (HPT) is a strategy to find the optimal set of hyper-parameters for training, and testing deep neural networks to achieve better convergence \cite{yu2020hyper, yang2020hyperparameter}. However, not much change has been done with a focus on hard samples to train networks for faster convergence.\\

\noindent \textbf{Data driven approaches.}
Data-driven approaches focus on the quality of the data rather than focus on model novelties \cite{lavecchia2015machine}. Some of the popular data-driven approaches are data augmentation \cite{shorten2019survey, mikolajczyk2018data}, feature engineering \cite{roh2019survey}, sampling \cite{johnson2019survey}, and data normalization \cite{singh2020investigating}. These approaches generally focus on improving the quality of the dataset, data transformation, and increasing the size of the dataset. To the authors' knowledge, none of the methodologies in the literature focus on dynamically selecting samples or a mini-batch of data for training.




\section{Method}
\label{sec:method}
The currently followed traditional approach for mini-batch training neural networks is defined by two hyper-parameters, the number of epochs $E$ and the batch size $\mathcal{B}$. The number of epochs $E$ is defined as the total number of times the network will go through the whole dataset. The batch-size $\mathcal{B}$ is the number of samples from the dataset to be propagated to the network (in mini-batches) in every iteration. For the process of training a network, a training dataset $\mathcal{D}_{T}$ is used, and for the evaluation of the learned network a test dataset $\mathcal{D}_{t}$ is used. In the case of standard benchmark datasets, the distribution of $\mathcal{D}_{T}$ and $\mathcal{D}_{t}$ is assumed to be similar. $\mathcal{D}_T$ contains $N$ mini-batches each of size $\mathcal{B}$ and $\mathcal{D}_t$ contains $M$ mini-batches of the same batch-size $\mathcal{B}$. The datasets are represented as $\mathcal{D}_{T} = \{(x_i, y_i)\}_{i = 0}^{N-1}$ and $\mathcal{D}_{t} = \{(x_i, y_i)\}_{i = 0}^{M-1}$, where $x$ denotes a mini-batch of images and $y$ denotes a mini-batch of labels, both of size $\mathcal{B}$. A single iteration corresponds to processing one mini-batch of samples.

\begin{figure*}[t]
    \centering
    \begin{tikzpicture}
[align=center, inner sep=0.8mm,thick,scale=0.62,
batch/.style={circle,draw=blue!50,fill=blue!20,thick, inner sep=0pt,minimum size=9mm}, 
model/.style={rectangle, draw=orange!50, fill=orange!20, thick, inner sep=5pt, minimum height=1.5cm,minimum width=2cm},
batch_pair/.style={rectangle split, rectangle split horizontal, rectangle split parts=2, draw=red!80, fill=red!20, thick, inner sep=4pt,minimum width=6pt, minimum height=7.2mm},
sorting/.style={rectangle, draw=yellow!80, fill=yellow!40, thick, inner sep=5pt, minimum height=1.5cm,minimum width=1cm},
delta_batch/.style={rectangle split, rectangle split horizontal, rectangle split parts=2, draw=green!70, fill=green!30, thick, inner sep=4pt,minimum width=6pt, minimum height=7.2mm},
every node/.style={transform shape}
]
\draw [decorate,decoration={brace,amplitude=10pt,mirror},xshift=-20pt] (-2.2,4.25) -- (-2.2,-4.25) 
node [black,midway,xshift=-0.8cm,rotate=-90] (n_batches) {$N$ mini-batches};

\draw [gray, densely dashed] (-1.8, 4) -- (0, 0) {};
\draw [gray, densely dashed] (-1.8, 3) -- (0, 0) {};
\draw [gray, densely dashed] (-1.8, 2) -- (0, 0) {};
\draw [gray, densely dashed] (-1.8, 1) -- (0, 0) {};
\draw [gray, densely dashed] (-1.8, 0) -- (0, 0) {};
\draw [gray, densely dashed] (-1.8, -1) -- (0, 0) {};
\draw [gray, densely dashed] (-1.8, -2) -- (0, 0) {};
\draw [gray, densely dashed] (-1.8, -3) -- (0, 0) {};
\draw [gray, densely dashed] (-1.8, -4) -- (0, 0) {};

\node at ( -2.2,4) [batch] {$b_0$};
\node at ( -2.2,3) [batch] {$b_1$};
\node at ( -2.2,2) [batch] {$b_2$};
\node at ( -2.2,1) [batch] {$b_3$};
\node at ( -2.2,0) [batch] {$.$};
\node at ( -2.2,-1) [batch] {$.$};
\node at ( -2.2,-2) [batch] {$.$};
\node at ( -2.2,-3) [batch] {$b_{N-2}$};
\node at ( -2.2,-4) [batch] {$b_{N-1}$};

\node[] at (1.35,1.25) {Forward Pass};
\draw [->] (0.1, 1) -- (2.6, 1) {};
\node at (1.35, 0) [model] (network) {Neural Network};
\draw [<-] (0.1, -1) -- (2.6, -1) {};
\node[] at (1.35,-1.35) {Back Propogation};

\draw[] (2.8, 0) -- (3.25, 0) {};
\draw[] (3.25, 0) -- (3.25, -5) {};
\draw[] (3.25, -5) -- (-4.65, -5) {};
\draw[] (-4.65, -5) -- (-4.65, 0) {};
\draw[-latex] (-4.65, 0) -- (-4, 0) {};
\node[] at (-0.6, -5.5) {\Large Train for $E$ epochs $= (N \times E)$ iterations};

\node[thick] at (-1,5.5) {\textbf{Mini-batch training method}};

\draw [thin] (3.8, 5.8) -- (3.8, -5.8) {};

\draw [gray, densely dashed] (4.9, 4) -- (7.2, 0) {};
\draw [gray, densely dashed] (4.9, 3) -- (7.2, 0) {};
\draw [gray, densely dashed] (4.9, 2) -- (7.2, 0) {};
\draw [gray, densely dashed] (4.9, 1) -- (7.2, 0) {};
\draw [gray, densely dashed] (4.9, 0) -- (7.2, 0) {};
\draw [gray, densely dashed] (4.9, -1) -- (7.2, 0) {};
\draw [gray, densely dashed] (4.9, -2) -- (7.2, 0) {};
\draw [gray, densely dashed] (4.9, -3) -- (7.2, 0) {};
\draw [gray, densely dashed] (4.9, -4) -- (7.2, 0) {};

\node at ( 4.6,4) [batch] {$b_0$};
\node at ( 4.6,3) [batch] {$b_1$};
\node at ( 4.6,2) [batch] {$b_2$};
\node at ( 4.6,1) [batch] {$b_3$};
\node at ( 4.6,0) [batch] {$.$};
\node at ( 4.6,-1) [batch] {$.$};
\node at ( 4.6,-2) [batch] {$.$};
\node at ( 4.6,-3) [batch] {$b_{N-2}$};
\node at ( 4.6,-4) [batch] {$b_{N-1}$};

\node[] at (8.58,1.25) {Forward Pass};
\draw [->] (7.3, 1) -- (9.8, 1) {};
\node at (8.58, 0) [model] {Neural Network};
\draw [<-] (7.3, -1) -- (9.8, -1) {};
\node[] at (8.58, -1.35) {Back-Propagation};

\draw [gray, densely dashed] (11.7, 4) -- (10, 0) {};
\draw [gray, densely dashed] (11.7, 3) -- (10, 0) {};
\draw [gray, densely dashed] (11.7, 2) -- (10, 0) {};
\draw [gray, densely dashed] (11.7, 1) -- (10, 0) {};
\draw [gray, densely dashed] (11.7, 0) -- (10, 0) {};
\draw [gray, densely dashed] (11.7, -1) -- (10, 0) {};
\draw [gray, densely dashed] (11.7, -2) -- (10, 0) {};
\draw [gray, densely dashed] (11.7, -3) -- (10, 0) {};
\draw [gray, densely dashed] (11.7, -4) -- (10, 0) {};

\node at ( 12,4) [batch_pair] {\nodepart{one} $b_0$ \nodepart{two} $\mathcal{L}_0$};
\node at ( 12,3) [batch_pair] {\nodepart{one} $b_1$ \nodepart{two} $\mathcal{L}_1$};
\node at ( 12,2) [batch_pair] {\nodepart{one} $b_2$ \nodepart{two} $\mathcal{L}_3$};
\node at ( 12,1) [batch_pair] {\nodepart{one} $b_3$ \nodepart{two} $\mathcal{L}_3$};
\node at ( 12,0) [batch_pair] {\nodepart{one} $.$ \nodepart{two} $.$};
\node at ( 12,-1) [batch_pair] {\nodepart{one} $.$ \nodepart{two} $.$};
\node at ( 12,-2) [batch_pair] {\nodepart{one} $.$ \nodepart{two} $.$};
\node at ( 12,-3) [batch_pair] {\nodepart{one} $b_{N-2}$ \nodepart{two} $\mathcal{L}_{N-2}$};
\node at ( 12,-4) [batch_pair] {\nodepart{one} $b_{N-1}$ \nodepart{two} $\mathcal{L}_{N-1}$};

\draw [-latex, thick] (12.6, 0) -- (13.4, 0) {};
\node at (14.8, 0) [sorting,align=center] {Sorting\\ mini-batches\\ in desc. order\\ of loss ($\mathcal{L}_i$)};

\draw [-latex, thick] (16.2, 0) -- (17, 0) {};

\node[] at (8, -5.5) {\Large $N$ iterations};

\draw [gray, densely dashed] (19.1, 4) -- (20.5, 0) {};
\draw [gray, densely dashed] (19.1, 3) -- (20.5, 0) {};
\draw [gray, densely dashed] (19, 2) -- (20.5, 0) {};
\draw [gray, densely dashed] (19, 1) -- (20.5, 0) {};
\draw [gray, densely dashed] (19.1, 0) -- (20.5, 0) {};
\draw [gray, densely dashed] (19.1, -1) -- (20.5, 0) {};

\node at (18.5,4) [delta_batch] {\nodepart{one} $b'_0$ \nodepart{two} $\mathcal{L'}_0$};
\node at (18.5,3) [delta_batch] {\nodepart{one} $b'_1$ \nodepart{two} $\mathcal{L'}_1$};
\node at (18.5,2) [delta_batch] {\nodepart{one} $b'_2$ \nodepart{two} $\mathcal{L'}_2$};
\node at (18.5,1) [delta_batch] {\nodepart{one} $.$ \nodepart{two} $.$};
\node at (18.5,0) [delta_batch] {\nodepart{one} $b'_{\delta N - 2}$ \nodepart{two} $\mathcal{L'}_{\delta N - 2}$};
\node at (18.5,-1) [delta_batch] {\nodepart{one} $b'_{\delta N - 1}$ \nodepart{two} $\mathcal{L'}_{\delta N - 1}$};
\node at (18.5,-2) [batch_pair, opacity=0.5] {\nodepart{one} $.$ \nodepart{two} $.$};
\node at (18.5,-3) [batch_pair, opacity=0.5] {\nodepart{one} $b'_{N-2}$ \nodepart{two} $\mathcal{L'}_{N-2}$};
\node at (18.5,-4) [batch_pair, opacity=0.5] {\nodepart{one} $b'_{N-1}$ \nodepart{two} $\mathcal{L'}_{N-1}$};

\node[] at (21.86,1.25) {Forward Pass};
\draw [->] (20.5, 1) -- (23.1, 1) {};
\node at (21.86, 0) [model] {Neural Network};
\draw [<-] (20.5, -1) -- (23.1, -1) {};
\node[] at (21.86, -1.35) {Back Propogation};

\draw[] (23.3, 0) -- (23.5, 0) {};
\draw[] (23.5, 0) -- (23.5, -5) {};
\draw[] (23.5, -5) -- (14.8, -5) {};
\draw[->, -latex] (14.8, -5) -- (14.8, -1.25) {};
\node[] at (19.2, -5.5) {\Large Repeat \LARGE $\zeta$ \Large times \LARGE $=(\zeta \times \delta \times N)$ \Large iterations};

\node[thick] at (13.9,5.5) {\textbf{Proposed method}};

\matrix [draw=none, below, inner ysep=0.3cm] at (current bounding box.south) {
    \node [batch,minimum size=10pt,label=right:A single mini-batch] {}; &[1mm]
    \node [batch_pair,minimum size=2pt,label=right:A pair of mini-batch and it's loss ${(b_i, \mathcal{L}_i)}$] {}; &[2mm]
    \node [delta_batch,minimum size=2pt,label=right:Selected first $\delta N$ mini-batches] {}; &[2mm] \\
     \\
};

\end{tikzpicture}
    \caption{\textbf{An overview of the existing mini-batch training method \cite{ruder2016overview} (left) and the proposed method (right)}. In the existing method, $N$ mini-batches are trained iteratively for $E$ epochs, with no importance for the under-represented mini-batches. In the proposed method, $(\delta \times N)$ mini-batches with high loss are trained in iterations, equating to the same iterations as the traditional mini-batch method. $\zeta$ denotes the number of times we repeat the process of selecting $(\delta \times N)$ mini-batches after sorting in descending order of loss.}
    \label{fig:workflow}
\end{figure*}
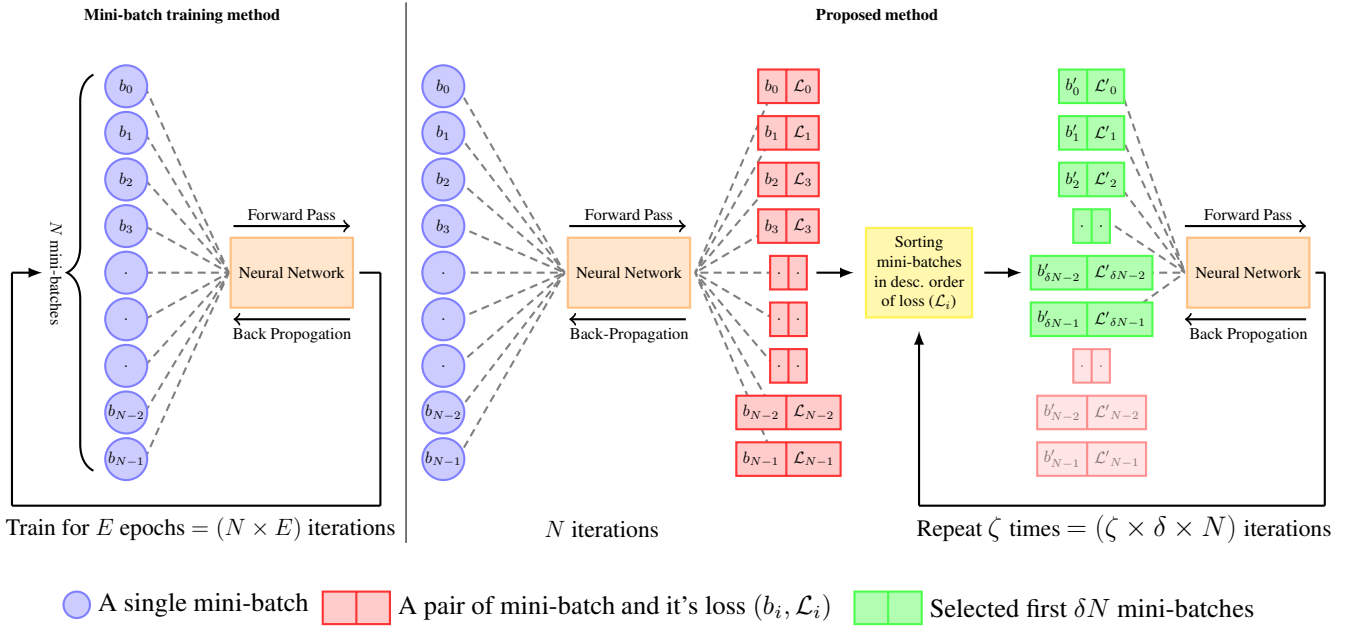

\subsection{Traditional Training Method}

Training in mini-batches SGD is the most common way of training a neural network. In mini-batch SGD \cite{bottou2010large,krizhevsky2017imagenet}, for every epoch, a total of $N$ mini-batches are propagated to the network in $N$ iterations. Specifically, in every iteration, one mini-batch of size $\mathcal{B}$ from the dataset $\mathcal{D}_{T}$ is passed to the network for back-propagation. 

\begin{algorithm}[tbh]
\DontPrintSemicolon
  \KwInput{Number of epochs $E$, Network $\mathcal{W}$ }
  \KwOutput{Trained weights $\mathcal{W}$}
  \KwData{$\mathcal{D}_{T} = \{(x_i, y_i)\}^{N-1}_{i=0}, \mathcal{D}_{t} = \{(x_i, y_i)\}^{M-1}_{i=0}$}
  \For{$e = 0, 1, \ldots, (E-1)$}{
    \tcc{Training $N$ mini-batches}
    \For{$(x_i, y_i) \in \mathcal{D}_T$}{
        \tcc{Forward Pass}
        $p = \mathcal{W}(x_i)$ \\
        \tcc{Calculate Train Loss}
        $\mathcal{L} \gets (y_i, p)$ \\
        \tcc{Back propagate loss on $\mathcal{W}$}
        $\mathcal{W} \gets \mathcal{L}$ \\
    }
    \tcc{Testing $M$ mini-batches}
    \For{$(x_i, y_i) \in \mathcal{D}_t$}{
        \tcc{Forward Pass}
        $p = \mathcal{W}(x_i)$ \\
        \tcc{Calculate Test Loss}
        $\mathcal{L} \gets (y_i, p)$ \\
    }
  }
  \Return{$\mathcal{W}$}
\caption{Traditional Training Approach}
\label{alg:standard}
\end{algorithm}

The loss function $\mathcal{L}$ is calculated over every mini-batch during the forward pass and then back-propagated for every mini-batch. The workflow is shown in Figure \ref{fig:workflow}, left panel. After the loss $\mathcal{L}$ is back-propagated and weights are updated for $N$ times for the training dataset, the resulting learned weights are used to validate the network on a test dataset $\mathcal{D}_t$. In the testing phase, the weights of the network don't change and are only used for the prediction of the $M$ mini-batches. The train and test metrics are averaged over the $M$ and $N$ mini-batches respectively.

The total number of back-propagations in Algorithm \ref{alg:standard} is $(N \times E)$ which is equal to the number of iterations and the number of forward passes.

\begin{equation}
\begin{split}
    \text{No. of training iterations} = N \times E \\
    \text{No. of back-propagations} = N \times E \\
    \text{No. of testing iterations} = M \times E \\
    \label{eq:standard}
\end{split}
\end{equation}

In an overview, the traditional mini-batch training method updates the weights $N \times E$ times, once in every epoch as given in Eq. \ref{eq:standard} equal to the total number of training iterations. 

\subsection{Proposed Training Method}

The proposed training approach focuses on learning the hard samples over the whole dataset through a new hyper-parameter $\delta$ which represents the fraction of the mini-batches to be considered for back-propagation. In the proposed approach, among the $N$ mini-batches in $\mathcal{D}_T$, only $\delta \times N$ mini-batches are selectively trained in each iteration. Since, $\delta \in (0, 1], (\delta \times N) \leq N, \forall N $ the model needs to train over the network for $(E - 1) / \delta$ times to ensure that the network is trained for the same number of weight updates. The number of times hard samples are trained within an epoch is called zeta ($\zeta$) as in Eq. \ref{eq:zeta}. The steps are shown in Algorithm \ref{alg:turbo}.

\begin{algorithm}[t]
\DontPrintSemicolon
  \KwInput{hyper-parameter $\delta$, Network $\mathcal{W}$, zeta $\zeta$}
  \KwOutput{Trained weights $\mathcal{W}$}
  \KwData{$\mathcal{D}_T = \{(x_i, y_i)\}^{N-1}_{i=0}, \mathcal{D}_t = \{(x_i, y_i)\}^{M-1}_{i=0}  $}
  \tcc{Compute loss for $N$ mini-batches}
  \tt{List = []} \\
  \For{$(x_i, y_i) \in \mathcal{D}^T$}{
        $p = \mathcal{W}(x_i)$ \\
        $\mathcal{L} \gets (y_i, p)$ \\
        \tcc{Store $\mathcal{L}$ for mini-batch i in $\mathcal{D^T}$}
        $\tt{List[i]} \gets (i, \mathcal{L})$ \\
        $\mathcal{W} \gets \mathcal{L}$
    }
    \tcc{Train $\zeta$ iterations with $\delta \times N$ mini-batches}
  \For{$z = 0, 1, \ldots, (\zeta - 1)$}{
  \tcc{Testing $M$ mini-batches}
    \For{$(x_i, y_i) \in \mathcal{D}_t$}{
        $p = \mathcal{W}(x_i)$ \\
        $\mathcal{L} \gets (y_i, p)$ \\
    }
  \tcc{Sort List in descending order of $\mathcal{L}$}
    Sorted(\tt{List}) \\
    \tcc{Train on first $\delta \times N$ mini-batches}
    $\mathcal{D} \gets \{(x_i, y_i)\}_{i=0}^{\delta \times N}$ \\
    \tcc{Size of $\mathcal{D} = \delta \times N$}
    \For{$(x_i, y_i) \in \mathcal{D}$}{
        $p = \mathcal{W}(x_i)$ \\
        $\mathcal{L} \gets (y_i, p)$ \\
        \tcc{Update the respective $\mathcal{L}$ in List}
        $\tt{List[i]} \gets (i, \mathcal{L})$ \\
        $\mathcal{W} \gets \mathcal{L}$
    }
  }
  \Return{$\mathcal{W}$}
\caption{Proposed Training Approach}
\label{alg:turbo}
\end{algorithm}

\begin{equation}
    \zeta = \frac{(E - 1)}{\delta}
    \label{eq:zeta}
\end{equation}

The network  is initially trained once on the $N$ mini-batches in the dataset to form a pair of the mini-batch $b_i$ and its corresponding loss $\mathcal{L}_i$ i.e; $(b_i, \mathcal{L}_i)$. These pairs are stored in a \texttt{List} of space complexity $O(N)$. The \texttt{List} can be represented as $\{(b_i, \mathcal{L}_i)\}_{i=0}^{N-1}$. The $\mathcal{L}_i$ in these pairs is updated for every back-propagation repeated $\zeta$ times followed 
by sorting. Sorting the \texttt{List} would incur an average time complexity of $O(N \times logN)$. 

The $N$ mini-batches are sorted in descending order of the loss $\mathcal{L}_i$. The order of the sorted mini-batch pairs is termed as $(b'_i, \mathcal{L}_i)$, where $b'_i$ is  the $i^{th}$ mini-batch in the $(\delta \times N)$ sorted mini-batches selected for training. The Loss $\mathcal{L}$ for these mini-batches is back-propagated to the network. This process is repeated $\zeta$ times as in Eq. \ref{eq:zeta} and \ref{eq:zeta_iter}.

\begin{equation}
    \text{\# of iterations for every $\zeta$} = (\zeta \times \delta \times N)
    \label{eq:zeta_iter}
\end{equation}

The number of times we are back-propagating in Algorithm \ref{alg:turbo} is
\begin{equation}
    \begin{split}
        \text{\# of back-propagations} & = N + \text{\# of iterations for every $\zeta$} \\
        & = N + (\zeta \times \delta \times N) \\
        & = N + (\frac{(E - 1)}{\delta} \times \delta \times N) \\
        & = N + (E - 1) \times N \\
        & = E \times N 
    \end{split}
\end{equation}

Thus, the total number of back-propagations in Algorithm \ref{alg:turbo} is $(N \times E)$ which is equal to the number of back-propagations in the standard algorithm \ref{alg:standard}. The proposed method focuses on the hardest $(\delta \times N)$ mini-batches every $\zeta$ number of times. Intuitively, the proposed method targets the hard samples in every dataset and trains them more to converge faster. The traditional mini-batch training method however, doesn't focus on training the under-represented samples in the dataset, which leads to more number of training iterations. 

\begin{table}[t]
    \centering
    \resizebox{0.5\textwidth}{!}
    {
    \begin{tabular}{ccccc}
        \hline
        \textbf{Method} & \textbf{\# of iter.} & \textbf{Time Complexity} & $\boldsymbol{\Delta t}$ (s) $\downarrow$ & $\boldsymbol{\Delta} (\boldsymbol{\Delta} t) \downarrow$\\
        \hline
        $\delta = 1.0$ & $(N \times E)$ & $O((N + M) \times E)$ & 0.0310 & -\\
        $\delta = 0.8$ & $(N \times E)$ & $O(N + \zeta (\delta \times N + M))$ & 0.0981 & \invdown{0.0671} \\
        $\delta = 0.5$ & $(N \times E)$ & $O(N + \zeta (\delta \times N + M))$ & 0.0991 & \invdown{0.0681} \\
        $\delta = 0.2$ & $(N \times E)$ & $O(N + \zeta (\delta \times N + M))$ & 0.0996 & \invdown{0.0686} \\
        \hline
    \end{tabular}
    }
    \caption{Comparision of differences in average time taken $(\Delta t)$ per iteration (i.e; per mini-batch) between different values of $\delta$ in ResNet-18 \cite{he2016deep} on CIFAR-10 \cite{krizhevsky2009learning}. $\Delta (\Delta t)$ denotes the change of $\Delta t$ with respect to $\delta=1$. \invdown{} denotes the change of time taken of traditional mini-batch training method with respect to the current $\delta$ ablation. The time complexity does not include the time taken for sorting $\texttt{List}$ which is $O(N \times logN)$.}
    \label{tab:runtime_comp}
\end{table}


\section{Experiments}
\label{sec:exps}

Image classification is a fundamental task when it comes to studying the performance of deep neural networks. To evaluate the performance of the proposed method, experiments have been conducted using well-known neural networks for image classification. To justify the performance of the proposed method, ablations are performed on a base hyper-parameter set and can be extended to any setting.

\subsection{Training Setup}
The codebase is built on the PyTorch \cite{paszke2019pytorch}, a machine learning framework, using \texttt{timm} deep learning library \cite{rw2019timm}, the standard for training classification models. All the experiments have been carried out on a Linux machine with a 40GB NVIDIA A100 GPU. To train the networks, the Loss function used is Cross Entropy Loss \cite{zhang2018generalized}, the optimizer SGD with momentum was preferred rather than Adam, as explained in the work \cite{zhou2020towards} with an initial learning rate of 0.005, a momentum of 0.9, and a mini-batch size of 512. The larger batch size is selected to efficiently utilize the available GPU RAM.

\subsection{Datasets}
The experiments have been conducted on three benchmark image classification datasets.   
The \textbf{CIFAR-10} \cite{krizhevsky2009learning} dataset consists of 60000 32x32 color images in 10 classes, with 6000 images per class. The standard training and testing splits have been used and they contain 50,000 and 10,000 samples, respectively. The \textbf{CIFAR-100} \cite{krizhevsky2009learning} dataset consists of 60000 32x32 color images in 100 classes, with 600 images per class. There are 50000 training images and 10000 test images. The dataset \textbf{STL-10} \cite{coates2011analysis} contains 5000 training images each of size 96x96 and 8000 testing images of the same size. These three benchmark datasets have been chosen to avoid data leaks and to ensure consistent results. Across all the datasets, the images have been cropped to the image size $128^2$.

\subsection{Evaluation Metrics}
The average top-1 accuracy at a 95\% confidence interval has been reported. The traditional mini-batch training has $(E \times N)$ iterations, while in the proposed training method, there are $N + (\zeta \times \delta \times N)$ iterations in total. So, we evaluate the metrics after the same number of back-propagations in both the traditional mini-batch method and the proposed method. So, the metrics for the traditional and the proposed methods are evaluated after every $N$ iterations and $1 / \delta$ respectively. We compare the generalization on the test top-1 accuracy. Similarly, we also compare networks on the basis of convergence on the train loss accordingly. Ablations are performed under simple settings to well understand the performance of compared networks and to show that the proposed method can be extended for various domain tasks.

\subsection{Results and Ablations}

\begin{table}
\resizebox{0.475\textwidth}{!}
{
\begin{tabular}{cccc}
\hline
\textbf{Network} & \textbf{image size} & \textbf{\# of params} & \textbf{FLOPs} \\ \hline
MobilenetV3-S \cite{howard2019searching}    & $128^2$    & 1.52 M       & 0.03 G  \\
EfficientNet-B4 \cite{tan2019efficientnet}      & $128^2$    & 17.56 M      & 0.98 G  \\
ResNet-18 \cite{he2016deep}             & $128^2$    & 11.18 M      & 1.19 G  \\
EfficientNetV2-S \cite{tan2021efficientnetv2} & $128^2$    & 20.19 M      & 1.86 G  \\
ResNet-50 \cite{he2016deep}             & $128^2$    & 23.52 M      & 2.69 G  \\
\hline
\end{tabular}
}
\caption{Comparison of \# of params and FLOPs across networks}
\label{tab:params_flops}
\end{table}

To evaluate the proposed method on well-known baseline networks, the authors have chosen five networks namely, ResNet-18, ResNet-50, Efficient Net B4, Efficient Net V2 Small, and Mobilenet V3. For effective comparison, the networks are selected to have a wider \# of parameters ranging from $1.52 M$ to $23.52 M$, tabulated in Table \ref{tab:params_flops}.

Tables \ref{tab:cifar10}, \ref{tab:cifar100}, and \ref{tab:stl10} show the performance comparison on the CIFAR-10, CIFAR-100, and STL-10 datasets between traditional mini-batch training ($\delta = 1$) and a proposed method with $\delta$ values of 0.2, 0.5, and 0.8 for several network architectures. The table includes the train and test top-1 accuracy in percentage, the epoch in which the network's train loss converges, and the percentage change ($\Delta e$) in the convergence epoch compared to $\delta = 1$. A positive $\Delta e$ value indicates an increase in the convergence speed, while a negative $\Delta e$ value indicates a decrease in the convergence speed.

\begin{table}
\resizebox{0.5\textwidth}{!}
{
\begin{tabular}{cccccc}
\hline
\textbf{Network} & $\boldsymbol{\delta}$ & \textbf{Train Top-1} (\%) $\uparrow$ & \textbf{Test Top-1} (\%) $\uparrow$ & $\boldsymbol{e}$ $\downarrow$ & $\boldsymbol{\Delta e}$ (\%) $\uparrow$ \\ \hline
\rowcolor[HTML]{C0C0C0} 
ResNet-18 & 1.0 & \val{99.7}{0.6} & \textbf{\val{69.6}{0.9}} & 80 & - \\
ResNet-18 & 0.8 & \val{99.7}{0.6} & \val{68.3}{1.0} & 79 & \up{1.26} \\
ResNet-18 & 0.5 & \val{99.7}{0.6} & \underline{\val{69.2}{1.0}} & 77 & \up{3.89} \\
ResNet-18 & 0.2 & \val{99.7}{0.7} & \val{68.7}{0.7} & 74 & \up{8.01} \\ \hline
ResNet-50 & 1.0 & \val{99.6}{0.7} & \val{63.1}{1.1} & 77 & -  \\
ResNet-50 & 0.8 & \val{99.6}{0.8} & \underline{\val{63.7}{0.9}} & 75 & \up{3.51} \\
\rowcolor[HTML]{C0C0C0} 
ResNet-50 & 0.5 & \val{99.6}{0.8} & \val{\textbf{64.6}}{\textbf{1.0}} & 72 & \up{6.94} \\
ResNet-50 & 0.2 & \val{99.6}{0.8} & \val{63.6}{1.0} & 70 & \up{10.00} \\ \hline
Efficient Net B4 & 1.0 & \val{99.5}{0.9} & \val{52.6}{0.7} & 34 & - \\
\rowcolor[HTML]{C0C0C0} 
Efficient Net B4 & 0.8 & \val{99.5}{0.9} & \textbf{\val{54.2}{1.0}} & 32 & \up{6.25} \\
Efficient Net B4 & 0.5 & \val{99.5}{1.0} & \val{49.7}{1.0} & 32 & \up{9.67} \\
Efficient Net B4 & 0.2 & \val{99.6}{0.9} & \underline{\val{54.0}{1.2}} & 30 & \up{13.33} \\ \hline
\rowcolor[HTML]{C0C0C0} 
EfficientNetV2-S & 1.0 & \val{99.6}{0.9} & \textbf{\val{56.5}{1.1}} & 33 & - \\
EfficientNetV2-S & 0.8 & \val{99.6}{0.9} & \underline{\val{53.8}{1.1}} & 27 & \up{22.22} \\
EfficientNetV2-S & 0.5 & \val{99.5}{0.9} & \val{53.1}{0.7} & 28 & \up{17.85} \\
EfficientNetV2-S & 0.2 & \val{99.5}{1.0} & \val{50.1}{1.3} & 24 & \up{27.21} \\ \hline
\rowcolor[HTML]{C0C0C0} 
MobilenetV3-S & 1.0 & \val{99.5}{0.9} & \textbf{\val{55.4}{1.1}} & 61 & - \\
MobilenetV3-S & 0.8 & \val{99.5}{1.0} & \val{51.7}{1.2} & 61 & \up{0.00} \\
MobilenetV3-S & 0.5 & \val{99.5}{1.0} & \val{51.7}{0.9} & 59 &\up{3.38} \\
MobilenetV3-S & 0.2 & \val{99.5}{0.9} & \underline{\val{53.2}{1.1}} & 53 & \up{15.09} \\ \hline
\end{tabular}
}
\caption{Performance comparison on \textbf{CIFAR-10} between the traditional mini-batch training ($\delta=1$) and the proposed method with $\delta=0.2,0.5,0.8$. $e$ is the epoch in which the training loss of the network converges. $\Delta e$ is the change between the $\delta=1$ and other compared $\delta$ values. \up{} denotes $+$ve change, \down{} denotes $-$ve change.}
\label{tab:cifar10}
\end{table}

Based on Table \ref{tab:cifar10}, it can be observed that the performance of the networks on CIFAR-10 varies depending on the network architecture and the value of $\delta$ used during training. In general, decreasing the value of $\delta$ leads to faster convergence and potentially similar generalization performance. The performance change can be observed in the test accuracy between $\delta = 1$ and the other values of $\delta$. For example, with ResNet-18, decreasing $\delta$ from 1.0 to 0.2 leads to an 8.01\% decrease in convergence time but only a 0.9\% decrease in test accuracy. However, this trend is not consistent across all networks, as decreasing $\delta$ from 1.0 to 0.2 actually leads to an increase in test accuracy for EfficientNet B4.
It is also worth noting that different network architectures have different performance characteristics, as shown by the differences in top-1 test accuracy and convergence time between the different networks. For example, EfficientNet B4 has the lowest top-1 test accuracy across all values of $\delta$, while ResNet-18 has the highest top-1 test accuracy for $\delta$ = 1.0 and 0.8.
Overall, the choice of network architecture and value of $\delta$ will depend on the specific application and trade-offs between training time and generalization performance.

\begin{table}
\resizebox{0.5\textwidth}{!}
{
\begin{tabular}{cccccc}
\hline
\textbf{Network} & $\boldsymbol{\delta}$ & \textbf{Train Top-1} (\%) $\uparrow$ & \textbf{Test Top-1} (\%) $\uparrow$ & $\boldsymbol{e}$ $\downarrow$ & $\boldsymbol{\Delta e}$ (\%) $\uparrow$ \\ \hline
ResNet-18 & 1.0 & \val{86.6}{1.4} & \val{33.2}{0.8} & 100 & - \\
ResNet-18 & 0.8 & \val{91.0}{1.4} & \val{33.6}{1.1} & 99 & \up{1.01} \\
ResNet-18 & 0.5 & \val{95.0}{1.3} & \underline{\val{34.5}{1.0}} & 92 & \up{8.69} \\
\rowcolor[HTML]{C0C0C0} 
ResNet-18 & 0.2 & \val{98.0}{1.3} & \textbf{\val{35.8}{0.9}} & 86 & \up{16.20} \\ \hline
ResNet-50 & 1.0 & \val{77.9}{1.5} & \val{32.9}{1.0} & 100 & - \\
ResNet-50 & 0.8 & \val{80.7}{1.4} & \val{33.2}{0.9} & 97 & \up{3.09} \\
\rowcolor[HTML]{C0C0C0} 
ResNet-50 & 0.5 & \val{88.3}{1.4} & \textbf{\val{35.0}{0.8}} & 93 & \up{7.52} \\
ResNet-50 & 0.2 & \val{95.1}{1.3} & \underline{\val{34.6}{1.0}} & 88 & \up{13.60} \\ \hline
Efficient Net B4 & 1.0 & \val{99.5}{0.9} & \val{52.6}{0.7} & 35 & - \\
\rowcolor[HTML]{C0C0C0} 
Efficient Net B4 & 0.8 & \val{99.5}{0.9} & \textbf{\val{54.2}{1.0}} & 35 & \up{0.00} \\
Efficient Net B4 & 0.5 & \val{99.5}{1.0} & \val{49.7}{1.0} & 42 & \down{16.6} \\
Efficient Net B4 & 0.2 & \val{99.6}{0.9} & \underline{\val{54.0}{1.2}} & 29 & \up{20.6} \\ \hline
\rowcolor[HTML]{C0C0C0} 
EfficientNetV2-S & 1.0 & \val{99.6}{0.9} & \textbf{\val{56.5}{1.1}} & 39 & - \\
EfficientNetV2-S & 0.8 & \val{99.6}{0.9} & \underline{\val{53.8}{1.1}} & 38 & \up{2.63} \\
EfficientNetV2-S & 0.5 & \val{99.5}{0.9} & \val{53.1}{0.7} & 37 & \up{5.41} \\
EfficientNetV2-S & 0.2 & \val{99.5}{1.0} & \val{50.1}{1.3} & 36 & \up{8.33} \\ \hline
\rowcolor[HTML]{C0C0C0} 
MobilenetV3-S & 1.0 & \val{70.3}{2.2} & \textbf{\val{21.5}{0.9}} & 95 & - \\
MobilenetV3-S & 0.8 & \val{81.2}{2.2} & \underline{\val{21.3}{0.8}} & 94 & \up{1.06} \\
MobilenetV3-S & 0.5 & \val{79.4}{2.7} & \val{19.7}{0.7} & 92 & \up{3.26} \\
MobilenetV3-S & 0.2 & \val{98.9}{1.7} & \val{19.3}{0.6} & 68 & \up{39.7} \\ \hline
\end{tabular}
}
\caption{Performance comparison on \textbf{CIFAR-100} between the traditional mini-batch training ($\delta=1$) and the proposed method with $\delta=0.2,0.5,0.8$. $e$ is the epoch in which the training loss of the network converges. $\Delta e$ is the change between the $\delta=1$ and other compared $\delta$ values. \up{} denotes $+$ve change, \down{} denotes $-$ve change.}
\label{tab:cifar100}
\end{table}

\par For CIFAR-100, the results in Table \ref{tab:cifar100} show that the proposed method with smaller $\delta$ values (0.2 and 0.5) results in better test top-1 accuracy and faster convergence compared to the traditional mini-batch training. For all the networks, decreasing the $\delta$ value to 0.2 resulted in a faster convergence epoch. On the other hand, larger networks like EfficientNet and Mobilenet show a considerable improvement in the time taken for convergence. Mobilenet appears to perform poorly compared to the other networks. It could be due to the smaller number of parameters in the network.

\begin{table}
\resizebox{0.5\textwidth}{!}
{
\begin{tabular}{cccccc}
\hline
\textbf{Network} & $\boldsymbol{\delta}$ & \textbf{Train Top-1} (\%) $\uparrow$ & \textbf{Test Top-1} (\%) $\uparrow$ & $\boldsymbol{e}$ $\downarrow$ & $\boldsymbol{\Delta e}$ (\%) $\uparrow$ \\ \hline
\rowcolor[HTML]{C0C0C0} 
ResNet-18 & 1.0 & \val{66.8}{4.5} & \textbf{\val{50}{1.5}} & 100 & - \\
ResNet-18 & 0.8 & \val{65.9}{4.6} & \val{48.3}{1.7} & 91 & \up{9.00} \\
ResNet-18 & 0.5 & \val{65.9}{4.0} & \underline{\val{49.5}{1.6}} & 91 & \up{9.00} \\
ResNet-18 & 0.2 & \val{60.9}{4.0} & \val{47.8}{1.0} & 97 & \up{3.09} \\ \hline

ResNet-50 & 1.0 & \val{82.1}{9.6} & \underline{\val{47.2}{1.4}} & 99 & - \\
ResNet-50 & 0.8 & \val{85.2}{9.7} & \val{47.1}{1.5} & 98 & \up{2.00} \\
\rowcolor[HTML]{C0C0C0} 
ResNet-50 & 0.5 & \val{85.1}{9.5} & \textbf{\val{47.3}{1.4}} & 97 & \up{3.00} \\
ResNet-50 & 0.2 & \val{77.1}{8.3} & \val{46.5}{1.5} & 100 & \up{0.00} \\  \hline

Efficient Net B4 & 1.0 & \val{93.0}{15.8} & \val{33.2}{0.9} & 43 & - \\
Efficient Net B4 & 0.8 & \val{93.1}{15.6} & \underline{\val{32.3}{1.1}} & 47 & \down{8.51} \\
Efficient Net B4 & 0.5 & \val{93.0}{15.9} & \val{31.0}{1.1} & 34 & \up{26.47} \\
\rowcolor[HTML]{C0C0C0} 
Efficient Net B4 & 0.2 & \val{92.7}{16.4} & \textbf{\val{33.5}{1.3}} & 41 & \up{7.50} \\ \hline

EfficientNetV2-S & 1.0 & \val{93.0}{15.8} & \val{37.6}{1.2} & 51 & - \\
EfficientNetV2-S & 0.8 & \val{93.4}{14.9} & \val{37.3}{1.1} & 58 & \down{12.06} \\
\rowcolor[HTML]{C0C0C0} 
EfficientNetV2-S & 0.5 & \val{93.9}{13.7} & \textbf{\val{39.9}{0.9}} & 44 & \up{15.90} \\
EfficientNetV2-S & 0.2 & \val{93.4}{14.8} & \underline{\val{38.2}{1.1}} & 54 & \down{5.50} \\ \hline

\rowcolor[HTML]{C0C0C0} 
MobilenetV3-S & 1.0 & \val{92.8}{16.2} & \textbf{\val{33.2}{1.0}} & 82 & - \\
MobilenetV3-S & 0.8 & \val{93.0}{15.9} & \val{32.6}{0.7} & 80 & \up{2.50} \\
MobilenetV3-S & 0.5 & \val{92.9}{16.2} & \underline{\val{32.8}{1.4}} & 83 & \down{1.20} \\
MobilenetV3-S & 0.2 & \val{92.8}{16.2} & \val{29.9}{1.1} & 81 & \down{1.23} \\ \hline
\end{tabular}
}
\caption{Performance comparison on \textbf{STL-10} between the traditional mini-batch training ($\delta=1$) and the proposed method with $\delta=0.2,0.5,0.8$. $e$ is the epoch in which the training loss of the network converges. $\Delta e$ is the change between the $\delta=1$ and other compared $\delta$ values. \up{} denotes $+$ve change, \down{} denotes $-$ve change.}
\label{tab:stl10}
\end{table}

\par Table \ref{tab:stl10} shows the results obtained on the STL-10 dataset. The model does converge faster for the ResNet models and Efficient Net B4. It can be noted that $\delta = 0.5$ converges faster compared to other values of $\delta$. From CIFAR-10 and CIFAR-100 results, the convergence has been fastest for $\delta  = 0.2$. However, it can be observed that for STL-10, $\delta = 0.5$ converges faster than $\delta = 0.2$. The authors conjecture that it could be due to the smaller STL-10 dataset size resulting in a smaller number of mini-batches. It can also be noted that the results on Mobilenet are marginal. 

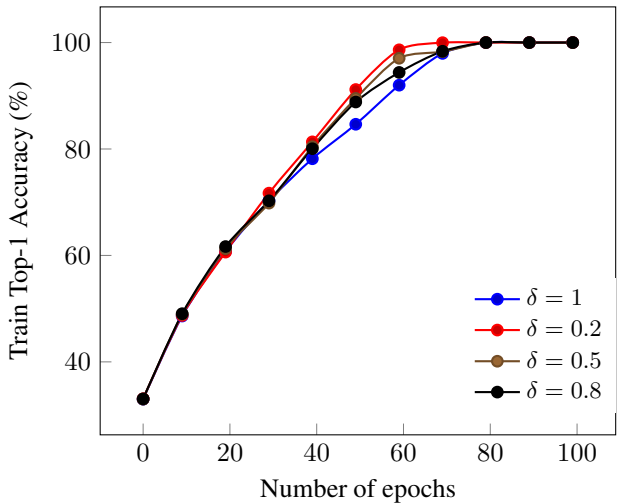
\begin{figure}[t]
    \centering
    \begin{tikzpicture}
    \begin{axis}[myplotstyle, xlabel=Number of epochs, ylabel=Train Top-1 Accuracy (\%), legend entries={$\delta = 1$, $\delta = 0.2$, $\delta = 0.5$, $\delta = 0.8$}, legend style={draw=none, at={(1, 0.05)} ,anchor=south east} ]
    \addplot+[smooth, mark=*] table [x=iter,y=delta_1.0_resnet18_cifar10_TrainTop-1Accuracy, col sep=comma] {runs_128.csv};
    \addplot+[smooth, mark=*] table [x=iter, y=delta_0.2_resnet18_cifar10_TrainTop-1Accuracy, col sep=comma] {runs_128.csv};
    \addplot+[smooth, mark=*] table [x=iter, y=delta_0.5_resnet18_cifar10_TrainTop-1Accuracy, col sep=comma] {runs_128.csv};
    \addplot+[smooth, mark=*] table [x=iter, y=delta_0.8_resnet18_cifar10_TrainTop-1Accuracy, col sep=comma] {runs_128.csv};
    \end{axis}
    \end{tikzpicture}    
    \caption{Generalization of ResNet-18 \cite{he2016deep} on different $\delta$ values on CIFAR-10 \cite{krizhevsky2009learning}. }
    \label{fig:r18_c10_train_acc}
\end{figure}

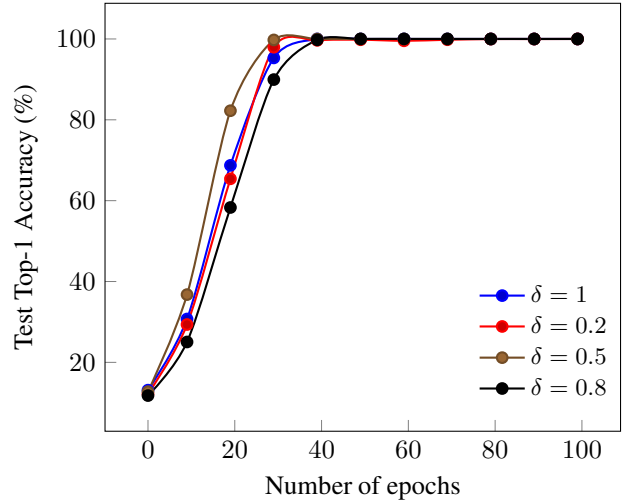
\begin{figure}[tbh]
    \centering
    \begin{tikzpicture}
    \begin{axis}[myplotstyle, xlabel=Number of epochs, ylabel=Test Top-1 Accuracy (\%), legend entries={$\delta = 1$, $\delta = 0.2$, $\delta = 0.5$, $\delta = 0.8$}, legend style={draw=none, at={(1, 0.05)} ,anchor=south east} ]
    \addplot+[smooth, mark=*] table [x=iter,y=delta_1.0_efficientnet_b4_stl10_TrainTop-1Accuracy, col sep=comma] {efb4_stl10.csv};
    \addplot+[smooth, mark=*] table [x=iter, y=delta_0.2_efficientnet_b4_stl10_TrainTop-1Accuracy, col sep=comma] {efb4_stl10.csv};
    \addplot+[smooth, mark=*] table [x=iter, y=delta_0.5_efficientnet_b4_stl10_TrainTop-1Accuracy, col sep=comma] {efb4_stl10.csv};
    \addplot+[smooth, mark=*] table [x=iter, y=delta_0.8_efficientnet_b4_stl10_TrainTop-1Accuracy, col sep=comma] {efb4_stl10.csv};
    \end{axis}
    \end{tikzpicture}    
    \caption{Generalization EfficientNet-B4 \cite{tan2019efficientnet} with different $\delta$ values on STL-10 \cite{coates2011analysis}}
    \label{fig:efb4_stl10_train_acc}
\end{figure}

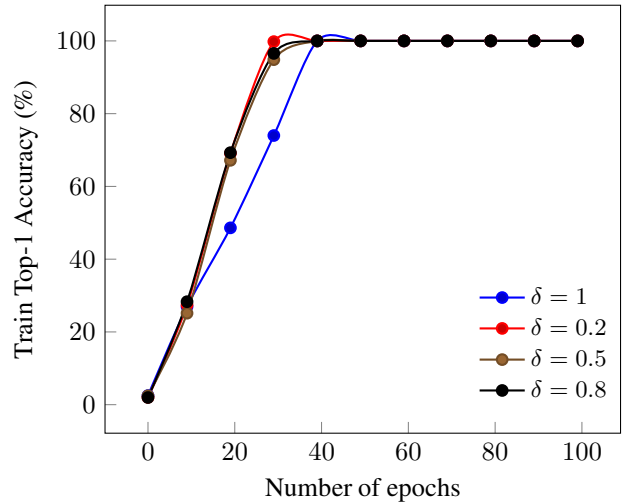
\begin{figure}[t]
    \centering
    \begin{tikzpicture}
    \begin{axis}[myplotstyle, xlabel=Number of epochs, ylabel=Train Top-1 Accuracy (\%), legend entries={$\delta = 1$, $\delta = 0.2$, $\delta = 0.5$, $\delta = 0.8$}, legend style={draw=none, at={(1, 0.05)} ,anchor=south east} ]
    \addplot+[smooth, mark=*] table [x=iter,y=delta_1.0_efficientnetv2_s_cifar100_TrainTop-1Accuracy, col sep=comma] {runs_128.csv};
    \addplot+[smooth, mark=*] table [x=iter, y=delta_0.2_efficientnetv2_s_cifar100_TrainTop-1Accuracy, col sep=comma] {runs_128.csv};
    \addplot+[smooth, mark=*] table [x=iter, y=delta_0.5_efficientnetv2_s_cifar100_TrainTop-1Accuracy, col sep=comma] {runs_128.csv};
   \addplot+[smooth, mark=*] table [x=iter, y=delta_0.8_efficientnetv2_s_cifar100_TrainTop-1Accuracy, col sep=comma] {runs_128.csv};
   \end{axis}
   \end{tikzpicture}    
   \caption{Generalization ResNet-18 \cite{he2016deep} with different $\delta$ values on CIFAR-100 \cite{krizhevsky2009learning} }
   \label{fig:efv2_cifar100_train_acc}
\end{figure}

Figures \ref{fig:r18_c10_train_acc}, \ref{fig:efb4_stl10_train_acc} illustrate the generalization of the ResNet-18 network on CIFAR-10 and CIFAR-100 on the train and test accuracy respectively for different values of $\delta$. In both the plots, it can be observed the proposed method with $\delta=0.2$ reaches generalization faster than all other values of $\delta$. 


\begin{table}[t]
\begin{tabular}{cccc}
\hline
& CIFAR-10 & CIFAR-100 & STL-10 \\ \hline
ResNet-18 & \downp{0.75} & \upp{7.26} & \downp{1.01} \\
ResNet-50 & \upp{2.32} & \upp{6.00} & \upp{0.21} \\
EfficientNet-B4 & \upp{2.95} & \upp{2.95}  & \upp{0.89} \\
EfficientNetV2-S & \downp{5.01} & \downp{5.01} & \upp{5.76} \\
MobileNetV3-S & \downp{4.13} & \downp{0.93} & \downp{1.21} \\ \hline
\end{tabular}
\caption{Overview of \textbf{Test Top-1 Acc.} in all networks across all datasets. For every comparison, the best ablation of $\delta$ is compared with $\delta=1.0$ from the Tables \ref{tab:cifar10}, \ref{tab:cifar100}, \ref{tab:stl10}. \up{} denotes improvement and \down{} denotes deterioration with respect to the traditional mini-batch training method (in \%).}
\label{tab:overview}
\end{table}

In Table \ref{tab:overview}, all the networks evaluated on all three datasets are compared with the best-generalized test accuracies with each other. It can be observed that the proposed method performs better in 8 out of 15 cases and with an average increase of 3.542 \% in the difference between traditional mini-batch training and the proposed method. The best-performing experiment is ResNet-18 on CIFAR-100 with an increase of 7.26 \%.

From the Tables \ref{tab:cifar10}, \ref{tab:cifar100}, \ref{tab:stl10}, it can be inferred that decreasing the value of $\delta$ in the proposed method can improve the test accuracy and speed up the convergence for most of the tested network architectures on different datasets. Smaller $\delta$ values (0.2 and 0.5) generally result in better test accuracy and faster convergence compared to the traditional mini-batch training ($\delta=1$). However, it can be noted that the choice of $\delta$ can depend on the total number of mini-batches as CIFAR-10 and CIFAR-100 has more mini-batches compared to STL-10. Consequently,  $\delta = 0.2$ converged faster for CIFAR-10 and CIFAR-100 datasets and $\delta = 0.5$ converged faster in STL-10 dataset. In addition, larger networks with more parameters like EfficientNet performed better than Mobilenet in the test results. Therefore, it is necessary to balance the choice of model selection, $\delta$ value according to the network architecture, and the dataset to achieve the best performance.

\section{Conclusion}
\label{sec:conclusion}
In conclusion, the study proposed a new method for mini-batch training that utilizes smaller batch sizes through the introduction of a new hyper-parameter $\delta$. The proposed method provides a new outlook to training neural networks with a focus on hard samples. The methodology is trained and validated over CIFAR-10, CIFAR-100, and STL-10 datasets. The networks used for the study are ResNet-18, ResNet-50, Efficient Net B4, EfficientNetV2-S, and MobilenetV3-S. The proposed methodology can be applied to any neural network training and can be extended across various tasks which involve back-propagation to improve generalization and faster convergence. Our findings suggest that the choice of $\delta$ value should be carefully balanced with respect to the network architecture, number of mini-batches, and dataset to achieve faster convergence and considerable performance.\\

\noindent Some of the limitations of the current approach can be briefed as follows, 
\begin{itemize}
    \item The work gives a new perspective to training deep learning networks using backpropagation. Though there are improvements when it comes to convergence, there aren't guarantees that the model can give increased performance. 
    \item The proposed method assumes the independence of the samples, which may not hold in some datasets, like in time series, 3D images, or videos.
    \item The proposed method has been studied for classification tasks only.
\end{itemize}

From the limitations, the future direction of the work will focus on improving the proposed algorithm, extending the work to include dependant data and explore applying the task for other tasks like object detection, segmentation, and so on.

{\small
\bibliographystyle{ieee_fullname}
\bibliography{output}

\begin{thebibliography}{10}\itemsep=-1pt

\bibitem{arpit2017closer}
Devansh Arpit, Stanis{\l}aw Jastrzebski, Nicolas Ballas, David Krueger,
  Emmanuel Bengio, Maxinder~S Kanwal, Tegan Maharaj, Asja Fischer, Aaron
  Courville, Yoshua Bengio, et~al.
\newblock A closer look at memorization in deep networks.
\newblock In {\em International conference on machine learning}, pages
  233--242. PMLR, 2017.

\bibitem{bengio2013representation}
Yoshua Bengio, Aaron Courville, and Pascal Vincent.
\newblock Representation learning: A review and new perspectives.
\newblock {\em IEEE transactions on pattern analysis and machine intelligence},
  35(8):1798--1828, 2013.

\bibitem{bengio2009learning}
Yoshua Bengio et~al.
\newblock Learning deep architectures for ai.
\newblock {\em Foundations and trends{\textregistered} in Machine Learning},
  2(1):1--127, 2009.

\bibitem{botev2017nesterov}
Aleksandar Botev, Guy Lever, and David Barber.
\newblock Nesterov's accelerated gradient and momentum as approximations to
  regularised update descent.
\newblock In {\em 2017 International joint conference on neural networks
  (IJCNN)}, pages 1899--1903. IEEE, 2017.

\bibitem{bottou2010large}
L{\'e}on Bottou.
\newblock Large-scale machine learning with stochastic gradient descent.
\newblock In {\em Proceedings of COMPSTAT'2010: 19th International Conference
  on Computational StatisticsParis France, August 22-27, 2010 Keynote, Invited
  and Contributed Papers}, pages 177--186. Springer, 2010.

\bibitem{chawla2010data}
Nitesh~V Chawla.
\newblock Data mining for imbalanced datasets: An overview.
\newblock {\em Data mining and knowledge discovery handbook}, pages 875--886,
  2010.

\bibitem{coates2011analysis}
Adam Coates, Andrew Ng, and Honglak Lee.
\newblock An analysis of single-layer networks in unsupervised feature
  learning.
\newblock In {\em Proceedings of the fourteenth international conference on
  artificial intelligence and statistics}, pages 215--223. JMLR Workshop and
  Conference Proceedings, 2011.

\bibitem{dogo2018comparative}
Eustace~M Dogo, OJ Afolabi, NI Nwulu, Bhekisipho Twala, and CO Aigbavboa.
\newblock A comparative analysis of gradient descent-based optimization
  algorithms on convolutional neural networks.
\newblock In {\em 2018 international conference on computational techniques,
  electronics and mechanical systems (CTEMS)}, pages 92--99. IEEE, 2018.

\bibitem{dosovitskiy2020image}
Alexey Dosovitskiy, Lucas Beyer, Alexander Kolesnikov, Dirk Weissenborn,
  Xiaohua Zhai, Thomas Unterthiner, Mostafa Dehghani, Matthias Minderer, Georg
  Heigold, Sylvain Gelly, et~al.
\newblock An image is worth 16x16 words: Transformers for image recognition at
  scale.
\newblock {\em arXiv preprint arXiv:2010.11929}, 2020.

\bibitem{fernandez2014we}
Manuel Fern{\'a}ndez-Delgado, Eva Cernadas, Sen{\'e}n Barro, and Dinani Amorim.
\newblock Do we need hundreds of classifiers to solve real world classification
  problems?
\newblock {\em The journal of machine learning research}, 15(1):3133--3181,
  2014.

\bibitem{frome2013devise}
Andrea Frome, Greg~S Corrado, Jon Shlens, Samy Bengio, Jeff Dean, Marc'Aurelio
  Ranzato, and Tomas Mikolov.
\newblock Devise: A deep visual-semantic embedding model.
\newblock {\em Advances in neural information processing systems}, 26, 2013.

\bibitem{geman1984stochastic}
Stuart Geman and Donald Geman.
\newblock Stochastic relaxation, gibbs distributions, and the bayesian
  restoration of images.
\newblock {\em IEEE Transactions on pattern analysis and machine intelligence},
  6(6):721--741, 1984.

\bibitem{guo2019spottune}
Yunhui Guo, Honghui Shi, Abhishek Kumar, Kristen Grauman, Tajana Rosing, and
  Rogerio Feris.
\newblock Spottune: transfer learning through adaptive fine-tuning.
\newblock In {\em Proceedings of the IEEE/CVF conference on computer vision and
  pattern recognition}, pages 4805--4814, 2019.

\bibitem{he2016deep}
Kaiming He, Xiangyu Zhang, Shaoqing Ren, and Jian Sun.
\newblock Deep residual learning for image recognition.
\newblock In {\em Proceedings of the IEEE conference on computer vision and
  pattern recognition}, pages 770--778, 2016.

\bibitem{hinton2006reducing}
Geoffrey~E Hinton and Ruslan~R Salakhutdinov.
\newblock Reducing the dimensionality of data with neural networks.
\newblock {\em science}, 313(5786):504--507, 2006.

\bibitem{howard2019searching}
Andrew Howard, Mark Sandler, Grace Chu, Liang-Chieh Chen, Bo Chen, Mingxing
  Tan, Weijun Wang, Yukun Zhu, Ruoming Pang, Vijay Vasudevan, et~al.
\newblock Searching for mobilenetv3.
\newblock In {\em Proceedings of the IEEE/CVF international conference on
  computer vision}, pages 1314--1324, 2019.

\bibitem{huang2017densely}
Gao Huang, Zhuang Liu, Laurens Van Der~Maaten, and Kilian~Q Weinberger.
\newblock Densely connected convolutional networks.
\newblock In {\em Proceedings of the IEEE conference on computer vision and
  pattern recognition}, pages 4700--4708, 2017.

\bibitem{iman2023review}
Mohammadreza Iman, Hamid~Reza Arabnia, and Khaled Rasheed.
\newblock A review of deep transfer learning and recent advancements.
\newblock {\em Technologies}, 11(2):40, 2023.

\bibitem{johnson2019survey}
Justin~M Johnson and Taghi~M Khoshgoftaar.
\newblock Survey on deep learning with class imbalance.
\newblock {\em Journal of Big Data}, 6(1):1--54, 2019.

\bibitem{kim2021automated}
Chiheon Kim, Saehoon Kim, Jongmin Kim, Donghoon Lee, and Sungwoong Kim.
\newblock Automated learning rate scheduler for large-batch training.
\newblock {\em arXiv preprint arXiv:2107.05855}, 2021.

\bibitem{kingma2014adam}
Diederik~P Kingma and Jimmy Ba.
\newblock Adam: A method for stochastic optimization.
\newblock {\em arXiv preprint arXiv:1412.6980}, 2014.

\bibitem{krawczyk2016learning}
Bartosz Krawczyk.
\newblock Learning from imbalanced data: open challenges and future directions.
\newblock {\em Progress in Artificial Intelligence}, 5(4):221--232, 2016.

\bibitem{krizhevsky2009learning}
Alex Krizhevsky, Geoffrey Hinton, et~al.
\newblock Learning multiple layers of features from tiny images.
\newblock {\em Master's thesis}, 2009.

\bibitem{krizhevsky2017imagenet}
Alex Krizhevsky, Ilya Sutskever, and Geoffrey~E Hinton.
\newblock Imagenet classification with deep convolutional neural networks.
\newblock {\em Communications of the ACM}, 60(6):84--90, 2017.

\bibitem{lavecchia2015machine}
Antonio Lavecchia.
\newblock Machine-learning approaches in drug discovery: methods and
  applications.
\newblock {\em Drug discovery today}, 20(3):318--331, 2015.

\bibitem{lecun2015deep}
Yann LeCun, Yoshua Bengio, and Geoffrey Hinton.
\newblock Deep learning.
\newblock {\em nature}, 521(7553):436--444, 2015.

\bibitem{lin2017focal}
Tsung-Yi Lin, Priya Goyal, Ross Girshick, Kaiming He, and Piotr Doll{\'a}r.
\newblock Focal loss for dense object detection.
\newblock In {\em Proceedings of the IEEE international conference on computer
  vision}, pages 2980--2988, 2017.

\bibitem{liu2021swin}
Ze Liu, Yutong Lin, Yue Cao, Han Hu, Yixuan Wei, Zheng Zhang, Stephen Lin, and
  Baining Guo.
\newblock Swin transformer: Hierarchical vision transformer using shifted
  windows.
\newblock In {\em Proceedings of the IEEE/CVF international conference on
  computer vision}, pages 10012--10022, 2021.

\bibitem{loshchilov2016sgdr}
Ilya Loshchilov and Frank Hutter.
\newblock Sgdr: Stochastic gradient descent with warm restarts.
\newblock {\em arXiv preprint arXiv:1608.03983}, 2016.

\bibitem{mikolajczyk2018data}
Agnieszka Miko{\l}ajczyk and Micha{\l} Grochowski.
\newblock Data augmentation for improving deep learning in image classification
  problem.
\newblock In {\em 2018 international interdisciplinary PhD workshop (IIPhDW)},
  pages 117--122. IEEE, 2018.

\bibitem{mukhoti2020calibrating}
Jishnu Mukhoti, Viveka Kulharia, Amartya Sanyal, Stuart Golodetz, Philip Torr,
  and Puneet Dokania.
\newblock Calibrating deep neural networks using focal loss.
\newblock {\em Advances in Neural Information Processing Systems},
  33:15288--15299, 2020.

\bibitem{neal2018modern}
Brady Neal, Sarthak Mittal, Aristide Baratin, Vinayak Tantia, Matthew Scicluna,
  Simon Lacoste-Julien, and Ioannis Mitliagkas.
\newblock A modern take on the bias-variance tradeoff in neural networks.
\newblock {\em arXiv preprint arXiv:1810.08591}, 2018.

\bibitem{neyshabur2020being}
Behnam Neyshabur, Hanie Sedghi, and Chiyuan Zhang.
\newblock What is being transferred in transfer learning?
\newblock {\em Advances in neural information processing systems}, 33:512--523,
  2020.

\bibitem{ozair2019wasserstein}
Sherjil Ozair, Corey Lynch, Yoshua Bengio, Aaron Van~den Oord, Sergey Levine,
  and Pierre Sermanet.
\newblock Wasserstein dependency measure for representation learning.
\newblock {\em Advances in Neural Information Processing Systems}, 32, 2019.

\bibitem{paszke2019pytorch}
Adam Paszke, Sam Gross, Francisco Massa, Adam Lerer, James Bradbury, Gregory
  Chanan, Trevor Killeen, Zeming Lin, Natalia Gimelshein, Luca Antiga, et~al.
\newblock Pytorch: An imperative style, high-performance deep learning library.
\newblock {\em Advances in neural information processing systems}, 32, 2019.

\bibitem{pruthi2020estimating}
Garima Pruthi, Frederick Liu, Satyen Kale, and Mukund Sundararajan.
\newblock Estimating training data influence by tracing gradient descent.
\newblock {\em Advances in Neural Information Processing Systems},
  33:19920--19930, 2020.

\bibitem{roh2019survey}
Yuji Roh, Geon Heo, and Steven~Euijong Whang.
\newblock A survey on data collection for machine learning: a big data-ai
  integration perspective.
\newblock {\em IEEE Transactions on Knowledge and Data Engineering},
  33(4):1328--1347, 2019.

\bibitem{rosenblatt1958perceptron}
Frank Rosenblatt.
\newblock The perceptron: a probabilistic model for information storage and
  organization in the brain.
\newblock {\em Psychological review}, 65(6):386, 1958.

\bibitem{ruder2016overview}
Sebastian Ruder.
\newblock An overview of gradient descent optimization algorithms.
\newblock {\em arXiv preprint arXiv:1609.04747}, 2016.

\bibitem{rumelhart1986learning}
David~E Rumelhart, Geoffrey~E Hinton, and Ronald~J Williams.
\newblock Learning representations by back-propagating errors.
\newblock {\em nature}, 323(6088):533--536, 1986.

\bibitem{shorten2019survey}
Connor Shorten and Taghi~M Khoshgoftaar.
\newblock A survey on image data augmentation for deep learning.
\newblock {\em Journal of big data}, 6(1):1--48, 2019.

\bibitem{simonyan2014very}
Karen Simonyan and Andrew Zisserman.
\newblock Very deep convolutional networks for large-scale image recognition.
\newblock {\em arXiv preprint arXiv:1409.1556}, 2014.

\bibitem{singh2020investigating}
Dalwinder Singh and Birmohan Singh.
\newblock Investigating the impact of data normalization on classification
  performance.
\newblock {\em Applied Soft Computing}, 97:105524, 2020.

\bibitem{staib2019escaping}
Matthew Staib, Sashank Reddi, Satyen Kale, Sanjiv Kumar, and Suvrit Sra.
\newblock Escaping saddle points with adaptive gradient methods.
\newblock In {\em International Conference on Machine Learning}, pages
  5956--5965. PMLR, 2019.

\bibitem{tan2019efficientnet}
Mingxing Tan and Quoc Le.
\newblock Efficientnet: Rethinking model scaling for convolutional neural
  networks.
\newblock In {\em International conference on machine learning}, pages
  6105--6114. PMLR, 2019.

\bibitem{tan2021efficientnetv2}
Mingxing Tan and Quoc Le.
\newblock Efficientnetv2: Smaller models and faster training.
\newblock In {\em International conference on machine learning}, pages
  10096--10106. PMLR, 2021.

\bibitem{twomey1998bias}
Janet~M Twomey and Alice~E Smith.
\newblock Bias and variance of validation methods for function approximation
  neural networks under conditions of sparse data.
\newblock {\em IEEE Transactions on Systems, Man, and Cybernetics, Part C
  (Applications and Reviews)}, 28(3):417--430, 1998.

\bibitem{werbos1974beyond}
Paul Werbos.
\newblock Beyond regression: New tools for prediction and analysis in the
  behavioral sciences.
\newblock {\em PhD thesis, Committee on Applied Mathematics, Harvard
  University, Cambridge, MA}, 1974.

\bibitem{rw2019timm}
Ross Wightman.
\newblock Pytorch image models.
\newblock \url{https://github.com/rwightman/pytorch-image-models}, 2019.

\bibitem{yang2020hyperparameter}
Li Yang and Abdallah Shami.
\newblock On hyperparameter optimization of machine learning algorithms: Theory
  and practice.
\newblock {\em Neurocomputing}, 415:295--316, 2020.

\bibitem{you2019large}
Yang You, Jing Li, Sashank Reddi, Jonathan Hseu, Sanjiv Kumar, Srinadh
  Bhojanapalli, Xiaodan Song, James Demmel, Kurt Keutzer, and Cho-Jui Hsieh.
\newblock Large batch optimization for deep learning: Training bert in 76
  minutes.
\newblock {\em arXiv preprint arXiv:1904.00962}, 2019.

\bibitem{yu2020hyper}
Tong Yu and Hong Zhu.
\newblock Hyper-parameter optimization: A review of algorithms and
  applications.
\newblock {\em arXiv preprint arXiv:2003.05689}, 2020.

\bibitem{zhang2019deep}
Shuai Zhang, Lina Yao, Aixin Sun, and Yi Tay.
\newblock Deep learning based recommender system: A survey and new
  perspectives.
\newblock {\em ACM computing surveys (CSUR)}, 52(1):1--38, 2019.

\bibitem{zhang2018generalized}
Zhilu Zhang and Mert Sabuncu.
\newblock Generalized cross entropy loss for training deep neural networks with
  noisy labels.
\newblock {\em Advances in neural information processing systems}, 31, 2018.

\bibitem{zhou2020towards}
Pan Zhou, Jiashi Feng, Chao Ma, Caiming Xiong, Steven Chu~Hong Hoi, et~al.
\newblock Towards theoretically understanding why sgd generalizes better than
  adam in deep learning.
\newblock {\em Advances in Neural Information Processing Systems},
  33:21285--21296, 2020.

\end{thebibliography}
}

\end{document}